\documentclass[preprint,12pt]{elsarticle}

\usepackage{amssymb}
\usepackage{subfigure}
\usepackage{array}
\usepackage{url}
\usepackage{amsmath,bm}
\usepackage{amsfonts}
\usepackage[pagebackref,breaklinks,colorlinks]{hyperref}

\usepackage{lineno}

\journal{Arxiv}

\begin{document}

\begin{frontmatter}



\title{Generalised Image Outpainting with U-Transformer}

\author[a]{Penglei Gao\fnref{1}}
\ead{P.Gao6@liverpool.ac.uk}
\author[b]{Xi Yang\fnref{1}}
\ead{xi.yang01@xjtlu.edu.cn}
\author[c]{Rui Zhang}
\author[a]{John Y. Goulermas}
\author[b]{Yujie Geng}
\author[b]{Yuyao Yan}
\author[d]{Kaizhu Huang\corref{*}}
\ead{kaizhu.huang@dukekunshan.edu.cn}

\cortext[*]{Corresponding author}
\fntext[1]{Authors contributed equally}

\address[a]{Department of Computer Science, University of Liverpool}
\address[b]{Department of Intelligent Science, Xi'an Jiaotong-Liverpool University}
\address[c]{Department of Foundational Mathematics, Xi'an Jiaotong-Liverpool University}
\address[d]{Institute of Applied Physical Sciences and Engineering, Duke Kunshan University}

\begin{abstract}
In this paper, we develop a novel transformer-based generative adversarial neural network called U-Transformer for generalised image outpainting problem. Different from most present image outpainting methods conducting horizontal extrapolation, our generalised image outpainting could extrapolate visual context all-side around a given image with plausible structure and details even for complicated scenery, building, and art images.
Specifically, we design a generator as an encoder-to-decoder structure embedded with the popular Swin Transformer blocks. As such, our novel neural network can better cope with image long-range dependencies which are crucially important for generalised image outpainting. We propose additionally a U-shaped structure and multi-view Temporal Spatial Predictor (TSP) module to reinforce image self-reconstruction as well as unknown-part prediction smoothly and realistically. By adjusting the predicting step in the TSP module in the testing stage, we can generate arbitrary outpainting size given the input sub-image. We experimentally demonstrate that our proposed method could produce visually appealing results for generalized image outpainting against the state-of-the-art image outpainting approaches.
\end{abstract}




\end{frontmatter}


\section{Introduction}
\begin{figure}[!htbp]
	\centering
	\subfigure[Horizontal extrapolation.]{
		\label{illust.1}
		\includegraphics[width=0.4\textwidth]{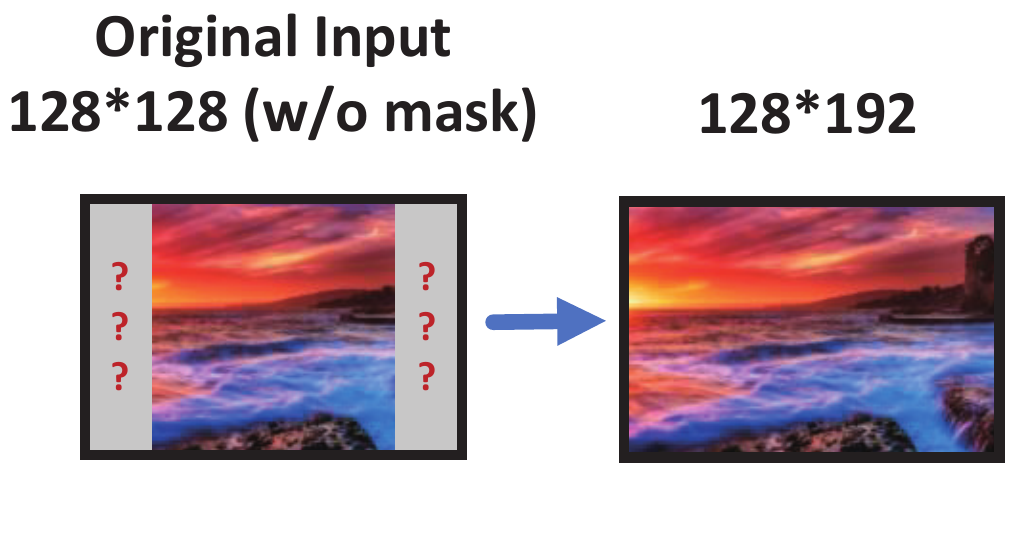}}
	\quad
	\subfigure[Generalised image outpainting.]{
		\label{illust.2}
		\includegraphics[width=0.39\textwidth]{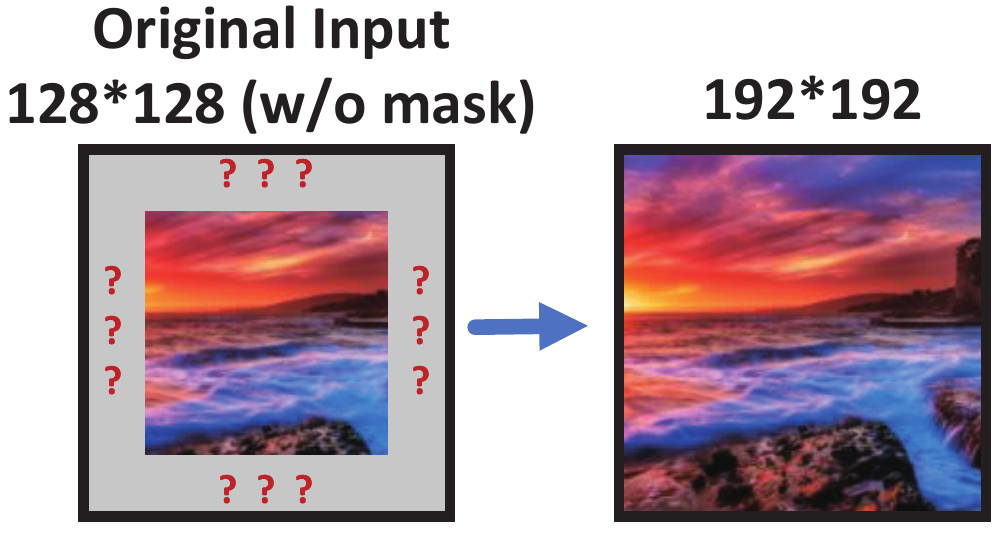}}
\caption{Illustration of image outpainting.}
	\label{illust}
\end{figure}

Reviewed from the previous work, most image outpainting neural networks conduct horizontal extrapolation, which expands the unknown-part image on one side. Differently, we study the generalised image outpainting problem as shown in Fig.~\ref{illust}, which extrapolates visual context all-side around a given image by leveraging the global modeling superiority of transformer-based networks.
Extrapolation on all sides has to handle the spatial relationship in both the horizontal and vertical directions (in Fig.~\ref{illust.2}) considering the size expansion and one-side constraints.
Unlike image inpainting \cite{pathak2016context,yu2018generative,yang2017high,FAN201834}, which makes use of rich contextual information, outpainting requires extrapolation to unknown image regions with less contextual information~\cite{kim2021painting}.
Dealing with such outpainting tasks, one has to predict and refill the missing regions with features absent from the input sub-image \cite{lin2021edge}. This leads to the challenges of designing neural networks of generating the exact and smooth scenes around the image for all boundaries as follows: 1) the missing features should be located relative to the spatial location of both near and distant features in the output; 2) the conditional input could be spatially distant from the missing regions to be predicted, which is dif?cult due to the lack of neighbouring ground truths.
Thus, the interesting and meaningful topic is to design an appropriate neural network to reconstruct and generate high-quality images that maintain the potential consistency in spatial configuration and semantic content with the original sub-images~\cite{yang2019very,lu2021bridging}.

\begin{figure}[!htbp]
	\centering
	\subfigure[]{
		\label{duibi.1}
		\includegraphics[width=0.2\textwidth]{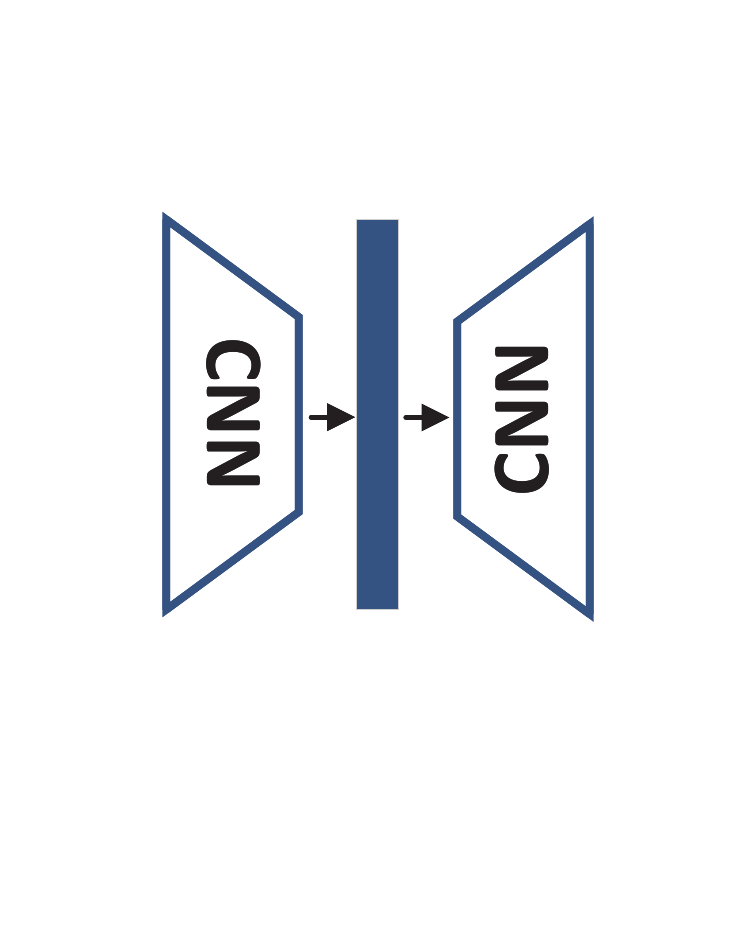}}
	\quad
	\subfigure[]{
		\label{duibi.2}
		\includegraphics[width=0.22\textwidth]{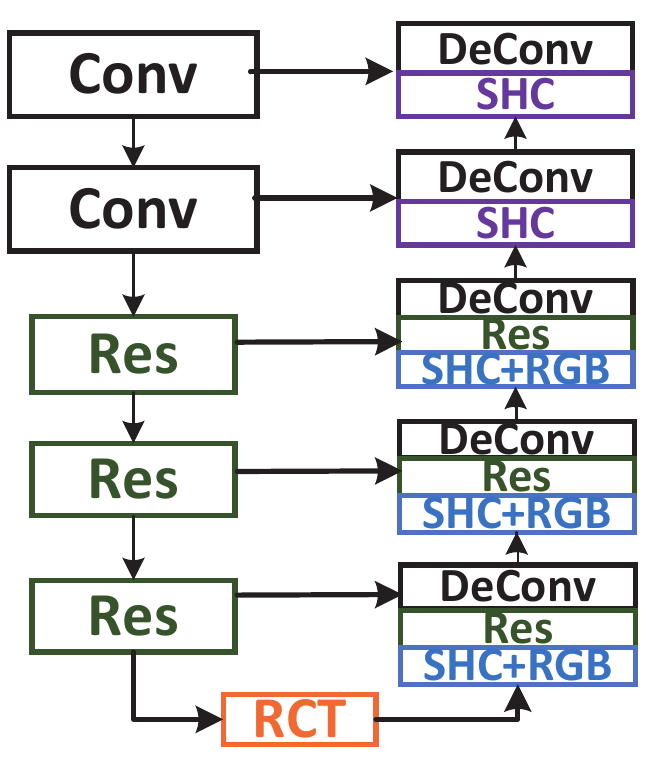}}
			\quad
	\subfigure[]{
		\label{duibi.3}
		\includegraphics[width=0.27\textwidth]{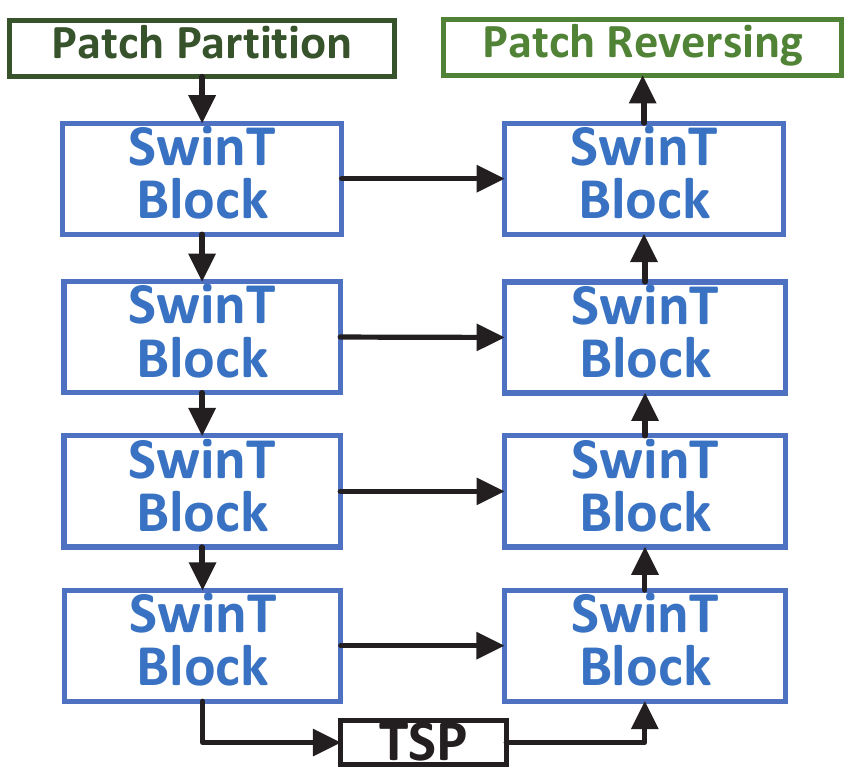}}
\caption{Image outpainting architectures. (a) Commonly used CNN-based Encoder-Decoder architecture; (b) U-Net architecture for Horizontal extrapolation task; (c) Proposed U-Transformer for Generalised image outpainting task.}
	\label{duibi}
\end{figure}

The existing horizontal and generalised image outpainting methods still suffer from blunt structures and abrupt colours when extrapolating the unknown regions of the images, although they have achieved solid performance~\cite{yang2019very,kim2021painting,wang2019wide, van2019image}.
The potential reason could be the fact that the generative adversarial network (GAN) Encoder-Decoder architecture (Fig.~\ref{duibi.1}) tends to ignore the spatial and semantic consistency only considering the semantic relationship between original and generated images. GAN and its variants are beneficial to the image generation performances in recent studies~\cite{ZHONG202019,SUN2020374}. The authors in \cite{yang2019very, lu2021bridging} exploited a U-Net structure to enhance the GAN's capability (Fig.~\ref{duibi.2}) allowing for the decoder to generate a more correlated prediction from the encoder involving the spatial relation.
As a matter of fact, all the existing deep learning-based image outpainting methods are based on convolutional neural networks (CNN) which have inductive biases \cite{d2021convit}. Such CNN backbone however suffers from the intrinsic locality and lacks the ability of capturing global features and representing long-range relations. The U-Net structure is limited in extracting the long-range semantic information interaction as well~\cite{chen2021transunet}.

To better cope with image long-range dependencies and spatial relationships among predicted regions and conditional input, we develop a novel transformer-based generative adversarial neural network called U-Transformer in this paper. Unlike present CNN-based outpainting networks, U-Transformer is the first model built on the transformer architecture able to extend image borders seamlessly. Concretely, we design our U-Transformer with the Swin Transformer~\cite{liu2021swin} blocks, which transform an image into a vector by using a flattened patch method and then encode the input sub-image with window self-attention (WSA). WSA controls the computation area by a window size, which makes the network easier to obtain global features for image long-range dependencies, especially for the predicted corner regions. We propose additionally a U-shaped structure and multi-view Temporal Spatial Predictor (TSP) to reinforce image self-reconstruction as well as unknown-part prediction  (Fig.~\ref{duibi}), both of which prove vital for generating image outpainting smoothly and realistically.
The evolution of images' details could be considered as changing with time in some aspects.
The tokenization of the embedded images is serialized and processed by the attention mechanism and recurrent networks to capture the potentially temporal information.

On the practical front, the transformer-based encoder and decoder with Swin Transformer blocks are used to extract context features.
The shifted window scheme has greater efficiency for non-overlapping patches computation in self-attention while  allowing for cross-window connection. Moreover, the U-shaped structure can enhance the spatial and semantic consistency for higher-quality image generation at the same layer level between encoder and decoder.
As for the bottleneck, the designed TSP module connects the encoder and decoder to capture the temporal information and spatial consistency among decomposed feature maps. Modelling the temporal relations and spatial correlations by a multi-view recurrent neural network (RNN) and self-attention block could help generate the outside regions smoothly and realistically.
Specifically, the output feature map of the encoder is divided into several small feature bars, which share a Long Short Term Memory (LSTM) \cite{hochreiter1997long} block. Then, each predicted feature bar is used to learn the spatial correlations via self-attention.
Besides, the width and height of the feature map control the size of the prediction region and the number of small feature bars. By adjusting the predicting step of the masked feature map in the TSP module, the proposed model supports the generation of arbitrary output resolutions.
Due to the fidelity expected at the higher resolution, Relativistic Average Least-Square GAN (RaLSGAN) \cite{Jolicoeur-Martineau19} is involved in making the generated images indistinguishable from the real images.

Our contributions are summarized in three-folds:
\begin{itemize}
\item To the best of our knowledge, the proposed U-Transformer is the first transformer-based image outpainting framework.
The Swin transformer blocks are able to obtain global features and keep high resolutions. A U-shaped structure and TSP module can reinforce image self-reconstruction as well as unknown-part prediction smoothly and realistically to enhance the network's capability.
\item The TSP module connects the encoder and decoder, which transfers incomplete latent features  considering the potentially temporal relations and spatial correlations by the multi-view LSTM network and self-attention blocks. Moreover, the TSP block is tractable to support the generation of arbitrary output resolutions by adjusting the predicting step of the masked feature map.
\item U-Transformer produces visually appealing results for generalised image outpainting against the state-of-the-art image outpainting networks. It also outperforms the competitors in most terms of quantitative results.
\end{itemize}

\section{Related Work}\label{sec:Rela}
\subsection{Image Outpainting}
Image outpainting fills the external regions of an image naturally and generates a high-quality larger natural image. Since the generated image usually resides beyond the boundaries of an original image and has insufficient contextual information, this task has two main challenges. One is to maintain the content and spatial of the generated region consistent with the original input. Another one is to generate high-quality larger images. Typically, image outpainting methods have been addressed only in a few studies on two derived tasks, horizontal outpainting and generalised outpainting~\cite{van2019image}.
Inspired by image inpainting methods, the image outpainting task was refocused by Mark Sabini and Gili Rusak~\cite{sabini2018painting} on a deep neural network framework. This effort focused on enhancing and smoothing the quality of generated images by using GANs and the post-processing methods to perform horizontal outpainting. Hoorick~\cite{van2019image} demonstrated that GAN has reasonable extrapolations to generate a similar content of the unknown region with the original image. Wang et al.~\cite{wang2019wide} proposed a Semantic Regeneration Network which learned semantic features directly from a small input image to avoid the bias in the general padding and up-sampling. However, these architectures tend to ignore the spatial and semantic consistency, so they still suffer from blunt structures and abrupt colours issues.
To solve this issue, Yang et al.~\cite{yang2019very} proposed a Recurrent Content Transfer (RCT) model for temporal content prediction by adopting a single LSTM model as the bottleneck. LSTM was designed to better control the horizontal information prediction for a very long-range image outpainting. To increase the contextual information, Lu et al.~\cite{lu2021bridging} and Kim et. al.~\cite{kim2021painting} rearranged the boundary region by switching the outer area of the image into its inner area. Moreover, Ma et al.~\cite{ma2021boosting} proposed a two-stage image outpainting framework that decomposed the task into two generation stages, semantic segmentation domain and image domain, to improve the quality and diversity of the output image.  However, these latest models are based on convolutional neural networks. Since global information is not well captured, they all have limitations in explicitly modelling long-range dependency.

\subsection{Transformer}
The transformer was first proposed to solve Natural Language Processing (NLP) tasks by discarding the traditional CNN and RNN structures. The entire network structure was composed  of Self-Attention and Feed Forward Neural Network entirely. Inspired by the success in the NLP field, efforts in the last two years made some adaptations of the Transformer architecture to make it suitable for basic visual feature extraction~\cite{vaswani2017attention}.
Self-attentive layers were used to replace some or all of the convolutional layers in ResNet, which achieved better accuracy and sped up the optimization~\cite{hu2019local, ramachandran2019stand}. However, the actual latency was significantly larger than that of convolutional networks because self-attention was computed within a local window of each pixel.
To better achieve the speed-accuracy trade-off on image classification, Vision Transformer (ViT)~\cite{dosovitskiy2020image} explored  a Transformer with global self-attention to full-size images directly.
Meanwhile, other variants of ViT also take efforts to improve the model performance on many vision tasks \cite{DeepViT,PiT,LeViT,BAKHTIARNIA2022461}.
Moreover, Liu et al~\cite{liu2021swin} proposed Swin Transformer to extend vision tasks to object detection and semantic segmentation. Swin Transformer involved Shifted window attention to bridge the windows of the preceding layer, which significantly enhanced modelling power as well as achieved lower latency.
In particular, Swin Transformer has been successfully applied in the field of medical image segmentation~\cite{cao2021swin} and salient object detection~\cite{liu2021swinnet}, achieving promising accuracy and showing strong feature representation ability.
To the best of our knowledge, the proposed U-Transformer is the first Transformer-based image outpainting framework.

\begin{figure}[t]
    \centering
    \includegraphics[width=0.9\textwidth]{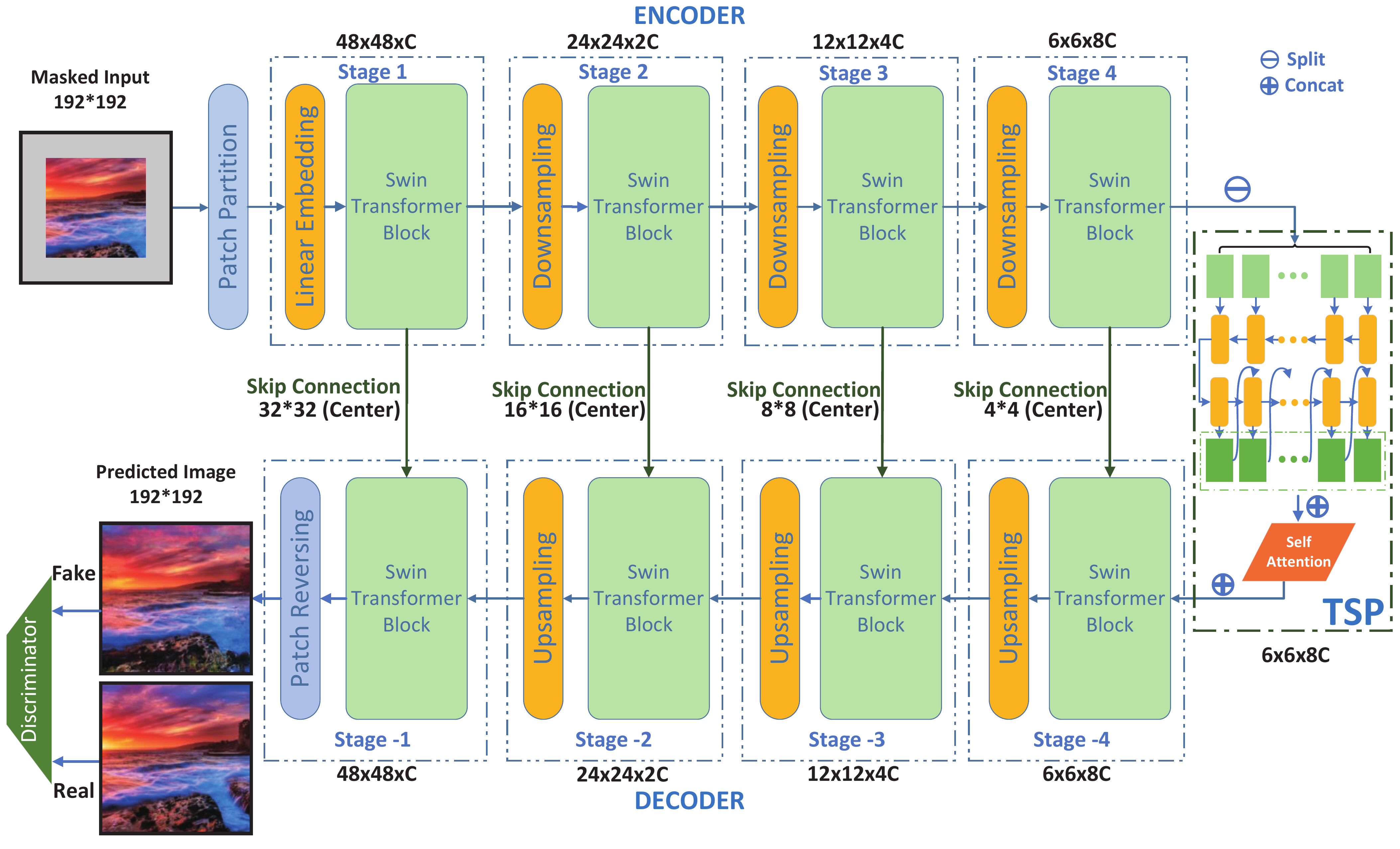}
    \caption{Overview of the U-Transformer for image outpainting consists of swin-transformer encoder-decoder, U-Net alike skip connection and temporal spatial predictor.}
    \label{struc}
\end{figure}
\section{Methodology}
The overall architecture is presented in Fig.~\ref{struc}. The underlying framework relies on a transformer-based encoder-decoder architecture, in which U-Net is utilized to enhance the self-reconstruction of the input images. In the feature prediction stage, we design a multi-view temporal and spatial information fusion approach.  The details of each component are described in the following sections.

\subsection{Problem Statement}
Given an image $I_{x} \in \mathbb{R}^{h\times w \times 3}$, we aim to extrapolate the outside contents beyond the image boundary with extra $m$-pixels.
The generator will produce a visually convincing image $I_{pred}\in \mathbb{R}^{h'\times w'\times 3}$, where $h'=h+2m,w'=w+2m$.
In practice, the outpainted image $I_{pred}$ is generated by mapping a partially masked RGB input image of size $h'\times w'$ with centre image $I_{x}$, which is denoted as $I_{mask}\in \mathbb{R}^{h'\times w'\times 3}$.  $I_{pred}$ has the same size as $I_{mask}$, in which the model's predictions replace the masked part.

\subsection{Overall Framework}\label{sec:GW}
Our proposed model is a U-Net alike encoder-decoder architecture for generalised image outpainting.
Generally, the basic encoder and decoder structures are derived from Swin Transformer~\cite{liu2021swin} by adding extra padding operation in the shifted window self-attention block for both encoder and decoder.

In the encoder $\mathcal{E}_{nc}$, the masked input images $I_{m}$ are first split into non-overlapping patches by a patch partition layer with a patch size of $4\times 4$.
The feature dimension of each patch is $4\times 4 \times 3 = 48$ since  each patch's feature could be considered a combination of the raw pixel RGB values. These patches features will be projected to latent features of dimension $C$ by a linear embedding layer.
After the patch embedding, there are $(\frac{H}{4}\times \frac{W}{4})$ patch tokens passed through several Swin Transformer blocks to obtain the hierarchical feature representations.
Moreover, a patch merging operation between two adjacent Swin Transformer blocks reduces the token resolution by $2\times$ down-sampling layers. We additionally apply a linear layer on the newly produced $4C^l$-dimensional features to convert the dimension to $2C^l$ in the $l$-th layer.

The decoder $\mathcal{D}_{ec}$ is composed of Swin Transformer blocks with an up-sampling layer and patch reversing block.
In contrast to a down-sampling layer, an up-sampling layer operates $2\times$ up-sampling resolution to expand the feature map. A linear layer is applied to convert the feature dimension of $2C^l$ to $4C^l$ before up-sampling at each layer level.
Inspired from U-Net \cite{ronneberger2019u}, we adopt Skip Connections (SC) to fuse and share the feature information from the encoder to the decoder at the same layer level (details in Section~\ref{sec:Unet}). The SC could encourage the decoder to learn the more accurate representation of centre reconstruction and generalized prediction taking full advantage of the information extracted from the encoder at each layer.

As for the bottleneck, we design a Temporal Spatial Predictor (TSP) module to propagate the information from input image feature maps to predicted feature maps which connect the encoder and decoder for more accurate outpainted features prediction. The detailed structure will be described in Section~\ref{sec:TSP}. In addition, the adversarial training is applied to encourage the generator $\mathcal{G}$ consisting of $\mathcal{E}_{nc}$, $TSP$, and $\mathcal{D}_{ec}$ to output more real images with a discriminator $\mathcal{D}_{is}$.

\subsection{Swin Transformer}\label{sec:STrans}
The global computation in standard self-attention makes it feasible to extract global features; however, it also leads to a huge complexity due to the increase in the number of tokens.
This makes it impracticable for image processing when representing high-resolution images.
For efficient modelling, we leverage Swin Transformer \cite{liu2021swin} which replaces the standard multi-head self-attention (MSA) module by utilizing shifted windows. The two successive Swin Transformer blocks are connected alternately, containing a window-based multi-head self-attention (W-MSA) module and a shifted-window-based multi-head self-attention (SW-MSA) module, respectively. The windows are arranged to evenly partition the image in a non-overlapping manner in the W-MSA module.
Then, the next SW-MSA module adopts a windowing configuration that is shifted from that of the preceding module. The shifted window partitioning approach captures the cross-relationship between neighbouring  non-overlapping windows in the previous layer.
Each block consists of a LayerNorm (LN) layer, a 2-layer MLP, and a GELU activation function with a residual connection between each module. According to the combined window partitioning approach, the consecutive Swin Transformer blocks are formulated as:
\begin{equation}
\begin{aligned}
&\hat{\bm{z}}^l= \text{W-MSA}(\text{LN}(\bm{z}^{l-1})) + \bm{z}^{l-1}, \\
&\bm{z}^l = \text{MLP}(\text{LN}(\hat{\bm{z}}^l)) + \hat{\bm{z}}^l,\\
&\hat{\bm{z}}^{l+1} = \text{SW-MSA}(\text{LN}(\bm{z}^l)) + \bm{z}^l,\\
&\bm{z}^{l+1} = \text{MLP}(\text{LN}(\hat{\bm{z}}^{l+1})) + \hat{\bm{z}}^{l+1},
\end{aligned}
\end{equation}
where $\hat{\bm{z}}^l$ and $\bm{z}^l$ are the output features of the (S)W-MSA module and the MLP module of block $l$ respectively.
Following the previous work \cite{hu2018relation,hu2019local,RaffelSRLNMZLL20}, the similarity of self-attention is computed as:
\begin{equation}
    \text{Att}(Q,K,V) = \text{SoftMax}(QK^\mathrm{T}/\sqrt{d} + B)V,
\end{equation}
where $Q,K,V\in \mathbb{R}^{M^2\times d}$ represent the $query$, $key$, and $value$ matrices and $B\in \mathbb{R}^{M^2\times M^2}$ is relative position bias. $M^2$ is the number of patches in a window and $d$ is the dimension of $query/key$. The bias $B$ is taken from a smaller-sized bias matrix $\hat{B}\in \mathbb{R}^{(2M-1)(2M-1)}$.

\subsection{Temporal Spatial Predictor}
\label{sec:TSP}
\begin{figure}[!htbp]
    \centering
    \includegraphics[width=0.5\textwidth]{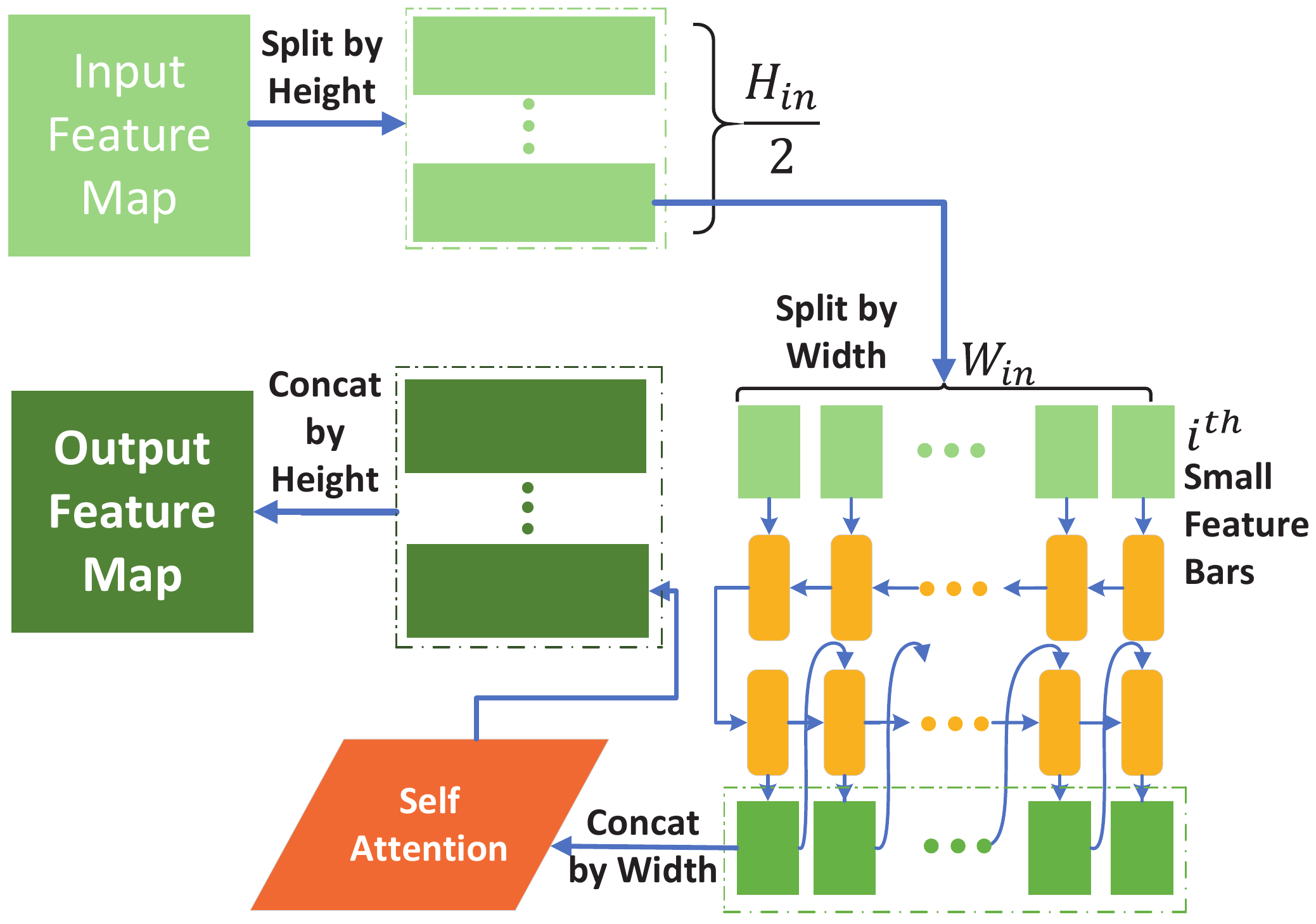}
    \caption{Temporal Spatial Predictor (TSP) structure.}
    \label{struc_TSP}
\end{figure}

The TSP module is designed for efficient information propagation and representation, considering the potentially temporal and spatial evolution in images.
The evolution of images' details could be considered as changing with time in some aspects.
The tokenization of the transformer encoder is serialized and could be operated by modeling the potentially temporal information.
As shown in Fig.~\ref{struc_TSP}, we vertically decompose the feature map from the encoder component into several small feature bars. Each small feature bar is split into a sequence in the width dimension and then processed by a 2-layer LSTM block to transfer this sequence to a new feature bar in the prediction region to capture the temporal information.  After that, the new feature bar and its neighbouring  feature bars as $value$ and $query/key$ will be processed by a standard multi-head self-attention to regulate the predicted features with the spatial information. In the end, the predicted feature bars are concatenated in the height dimension, forming the predicted feature map with a normalization layer. In TSP, the size of the prediction region is adapted by the width of the feature map in one-step predictions. The height of the feature map adapts to the number of small feature bars. According to the size of the masked feature map, the TSP module could output target feature size by iterating the LSTM block and self-attention block to generate images with high quality. By adjusting the predicting step in the testing stage, our model could support multi-step prediction to generate extrapolated images with arbitrary output resolutions.

\subsection{Skip Connection}
\label{sec:Unet}
To enhance the modelling ability of the decoder, we leverage skip connections between encoder and decoder. The un-masked centre feature maps at each encoder layer level are merged to the decoder layer level to share information. Given a feature from the $l$-$th$ layer in encoder $fe_{h^l,w^l,c^l}$ and a feature from the corresponding layer in decoder $fd_{h^l,w^l,c^l}$, SC computes a new feature $fd'_{h^l,w^l,c^l}$ as follows:
\begin{equation}
    fd'_{h^l_{cen},w^l_{cen},c^l} = (fe_{h^l_{cen},w^l_{cen},c^l} + fd_{h^l_{cen},w^l_{cen},c^l}) / 2.
\end{equation}
Then $fd'_{h^l,w^l,c^l} :=$ substitute the center part of $fd_{h^l,w^l,c^l}$ with $fd'_{h^l_{cen},w^l_{cen},c^l}$, where $h^l_{cen},w^l_{cen}$ represent the center size of a given feature map.

\subsection{Loss Function}
Our loss function consists of four parts: pixel reconstruction loss, feature reconstruction loss, texture consistency loss, and adversarial loss. The pixel reconstruction loss captures the logical coherence between the generated images and ground truth from the overall structure, which focuses on low-order information. The feature reconstruction loss is responsible for the enhancement of self-reconstruction in latent space. The texture consistency loss is utilized to encourage similar feature distribution between fake images and real images. The adversarial loss learns to make the predicted images more real due to high-order information capturing.

\textbf{Pixel Reconstruction Loss}
We use an L1 distance between ground truth image $I_{gt}$ and generated image $I_{pred}$ as the pixel reconstruction loss denoted as $\mathcal{L}_{rec}(X)$,
\begin{equation}
    \mathcal{L}_{rec}(I_{mask}) = ||I_{gt} - I_{pred}||_1.
\end{equation}

\textbf{Feature Reconstruction Loss}
When performing self-reconstruction, we could extract the feature map of the un-masked centre ground truth image $I_x$ via the encoder $\mathcal{E}_{nc}$, and encourage the predicted centre  feature from the TSP module to be identical to it. The feature reconstruction loss $\mathcal{L}_{feat\_rec}$ is defined as:
\begin{equation}
    \mathcal{L}_{feat\_rec} = ||f_{cen}-\mathcal{E}_{nc}(I_x)||_1,
\end{equation}
where $f_{cen}$ is the centre  part of the output from the TSP module.

\textbf{Texture Consistency Loss} According to previous works of impainting, outpainting, and image translation \cite{mechrez2018contextual,WangTQSJ18,wang2019wide}, we add the regularization of implicit diversified Markov random fields (IDMRF) to our loss function. IDMRF loss models the feature difference between $F(I_{fake})$ and $F(I_{real})$ and pushes $I_{fake}$ to have similar feature distribution as ground truth in self-reconstruction, where $F$ is a pretrained feature extractor. This IDMRF loss is represented as:
\begin{equation}
    \mathcal{L}_{mrf} = \text{IDMRF}(I_{fake}, I_{real}),
\end{equation}
where the computation of IDMRF is identical to the one in \cite{WangTQSJ18}. We use the layers $relu3_2$ and $relu4_2$ of a pretrained VGG19 \cite{simonyan2014very} network as the feature extractor $F$.

\textbf{Adversarial Loss}
The adversarial loss for balancing the discriminator $\mathcal{D}_{is}$ and the generator $\mathcal{G}$ is defined as follows:
\begin{eqnarray}
\begin{aligned}
    \mathcal{L}_{GAN}^{I_{gt}}=&\mathbb{E}_{I_{mask}\sim p(I_{mask})}[\log(1-\mathcal{D}_{is}(\mathcal{G}(I_{mask}))]   \\
    &+ \mathbb{E}_{I_{gt}\sim p(I_{gt})}[\log \mathcal{D}_{is}(I_{gt})].
\end{aligned}
\end{eqnarray}
In the experimental implementation, we leverage the RaLSGAN \cite{Jolicoeur-Martineau19} to perform adversarial learning for the discriminator because of its advantages of having more stable training and generating higher-quality images.


\textbf{Overall Objectives}
We jointly train the generator and discriminator to optimize the overall objective, which is a weighted sum of the aforementioned objectives:
\begin{eqnarray}
\begin{aligned}
    \mathcal{L} =& \min_{\mathcal{G}}\mathop{\max}_{\mathcal{D}_{is}}\mathcal{L}(\mathcal{G},\mathcal{D}_{is}) \\
    =& \lambda_{rec}\mathcal{L}_{rec} + \lambda_{feat\_rec}\mathcal{L}_{feat\_rec} \\
    &+ \lambda_{mrf}\mathcal{L}_{mrf} + \mathcal{L}_{GAN}^{I_{gt}}.
\end{aligned}
\end{eqnarray}
In our experiments, we set $\lambda_{rec}=20$, $\lambda_{feat\_rec}=1$, $\lambda_{mrf}=0.5$, and $\lambda_{adv}=1$.

\section{Experiments}
\subsection{Dataset}
In our experiments, we use the scenery dataset proposed by \cite{yang2019very} and the fine-art paintings dataset obtained from the wikiart.org website~\cite{wikiart} to conduct the generalised image outpainting for comparison. The scenery dataset consists of about 5,000 images for training and 1,000 images for testing. As for the wikiart dataset, there are about 45,500 training images and 19,500 testing images.
In addition, we prepare a new building facades dataset in different styles consisting of diverse and complicated building architecture. There are about 16,000 images in the training set and 1,500 images in the testing set. All the images are collected on the internet.
Our code and datasets could be found at \url{https://github.com/PengleiGao/UTransformer}.

\subsection{Implementation Details}

In our experimental setting, we first pre-process the data by resizing the original images to $192\times 192$ as ground truth images $I_{gt}$. The masked input images $I_{mask}$ are obtained by centre  cropping operation on $I_{gt}$ with size $128\times 128$ and expanding masks around the images with an extra $32$-pixels. The total output size is 2.25 times that of the input, indicating that over half of all pixels will be generated. For both encoder and decoder, the number of layers is 4, the corresponding depth is $[2,2,6,2]$, and the number of head is $[3,6,12,24]$ for each layer. The window size is set to $M=7$ and the query dimension of each head is $d=32$. The embedding dimension $C$ is 96.
The experiments are implemented with PyTorch 1.7 and trained on a single NVidia RTX3090.
We use Frechet Inception distance (FID) \cite{heusel2017gans}, Inception Score (IS) \cite{salimans2016improved}, peak signal-to-noise ratio (PSNR), and structural similarity index measure (SSIM) metrics for quantitative comparison.

\subsection{Experimental Results}
In the following, we first compare the proposed model with the previous state-of-the-art on generalized image outpainting. Then, we ablate the important designs of U-Transformer.
We choose three  baseline models for comparison with quantitative and qualitative evaluation. Model IOH \cite{van2019image} and SRN \cite{wang2019wide} are current generalised image outpainting methods. IOH is an encoder-decoder structure by using CNN with adversarial loss. SRN leverages two encoder-decoder structures for feature expansion and context prediction. NSIPO \cite{yang2019very} uses ResNet \cite{he2016deep} as the backbone and equipping  with skip horizontal connection (SHC) and recurrent content transfer (BCT) for horizontal extrapolation.

\subsubsection{Quantitative Results}
The quantitative results of evaluation metrics FID, IS, PSNR, SSIM against other state-of-the-art models on the scenery, building, and wikiart datasets are described in Table~\ref{quan}. Notably, our proposed U-Transformer achieves the best scores among most of the evaluation metrics, showing that our transformer-based network could model the global and distant relations for all-side extrapolations.
On the other hand, our model has slightly underperformed NSIPO on both FID and IS in the building dataset. The reason might be that the images in the building dataset are more complex to learn the overall structure. However, the extrapolated images generated by our model have vivid details and are more smooth in terms of visualization.
\begin{table}[!htbp]
\small
\caption{Quantitative results compared with other outpainting methods.}
\label{quan}
\centering
\scalebox{0.75}{
\begin{tabular}{|c|cccc|cccc|cccc|}
\hline
Dataset       & \multicolumn{4}{c|}{Scenery}                      & \multicolumn{4}{c|}{Building}                      & \multicolumn{4}{c|}{Wikiart}                       \\ \hline
Metrics       & FID$\downarrow$            & IS$\uparrow$             & PSNR$\uparrow$  &SSIM$\uparrow$          & FID$\downarrow$             & IS$\uparrow$             & PSNR$\uparrow$ &SSIM$\uparrow$           & FID$\downarrow$             & IS$\uparrow$             & PSNR$\uparrow$  &SSIM$\uparrow$          \\ \hline
SRN          & 62.781      & 3.048    & 22.421 & 0.747        & 47.097          & 4.072          & 18.614  & 0.676         & 96.693          & 3.641          & \textbf{19.921} & 0.644 \\
IOH           & 42.251         & 2.988          & 22.512  & 0.669        & 62.794          & 4.302          & 18.512   & 0.630       & 59.326          & 5.008          & 19.316  & 0.592        \\
NSIPO        & 34.542         & 3.380           & 21.250   & 0.755        & \textbf{37.905} & \textbf{4.513} & 18.398    & 0.703      & 33.937          & 6.412          & 18.571 & 0.677          \\
U-Transformer & \textbf{24.860} & \textbf{3.554} & \textbf{23.273} & \textbf{0.788} & 38.390           & 4.489          & \textbf{18.932} & \textbf{0.715} & \textbf{25.385} & \textbf{7.223} & 19.535    & \textbf{0.691}      \\ \hline
\end{tabular}}
\end{table}

\begin{table}[!htbp]
\caption{Comparison on inference time.}
\label{tab:time}
\centering
\begin{tabular}{c|cccc}
\hline
Method         & SRN    & NSIPO  & IOH   & Uformer \\ \hline
Time usage~(ms/image) & 11.960 & 44.190 & 4.160 & 46.810 \\ \hline
\end{tabular}
\end{table}

\begin{figure}[!htbp]
    \centering
    \includegraphics[width=0.95\textwidth]{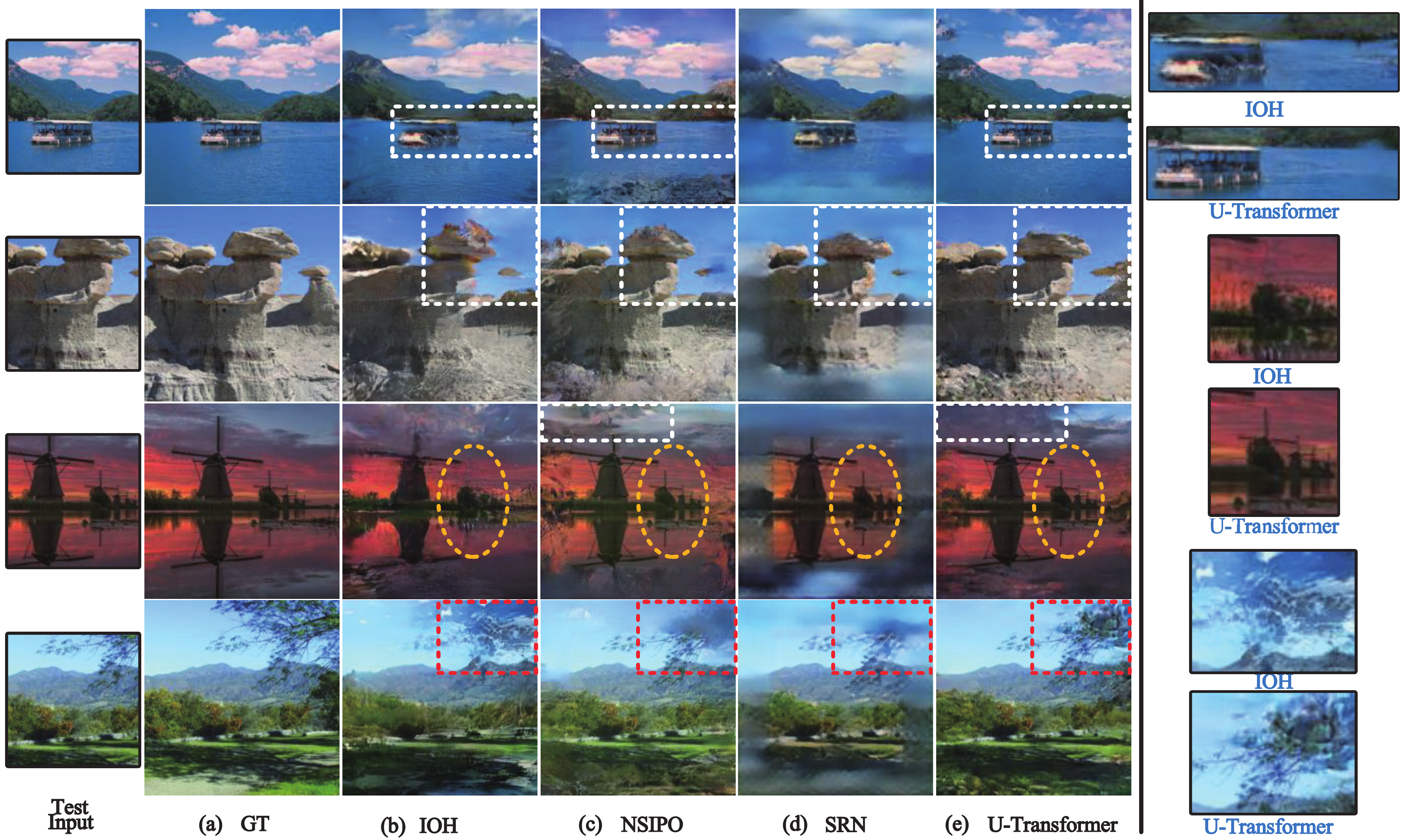}
    \caption{Qualitative results compared with other outpainting methods on scenery dataset.}
    \label{Quali}
\end{figure}

\begin{figure}[!htbp]
    \centering
    \includegraphics[width=0.95\textwidth]{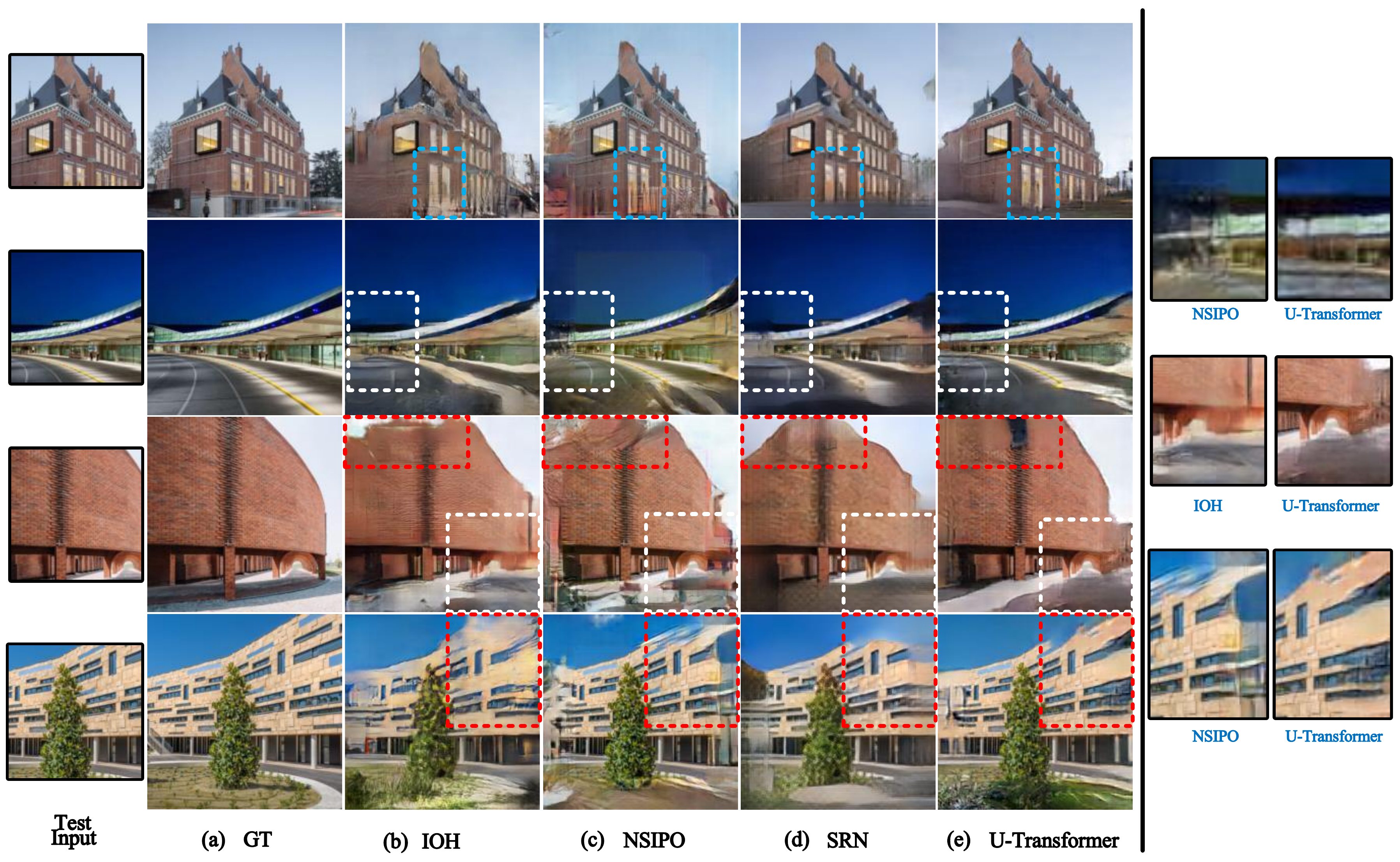}
    \caption{Qualitative results compared with other outpainting methods on building dataset.}
    \label{build}
\end{figure}

\begin{figure}[!htbp]
    \centering
    \includegraphics[width=0.95\textwidth]{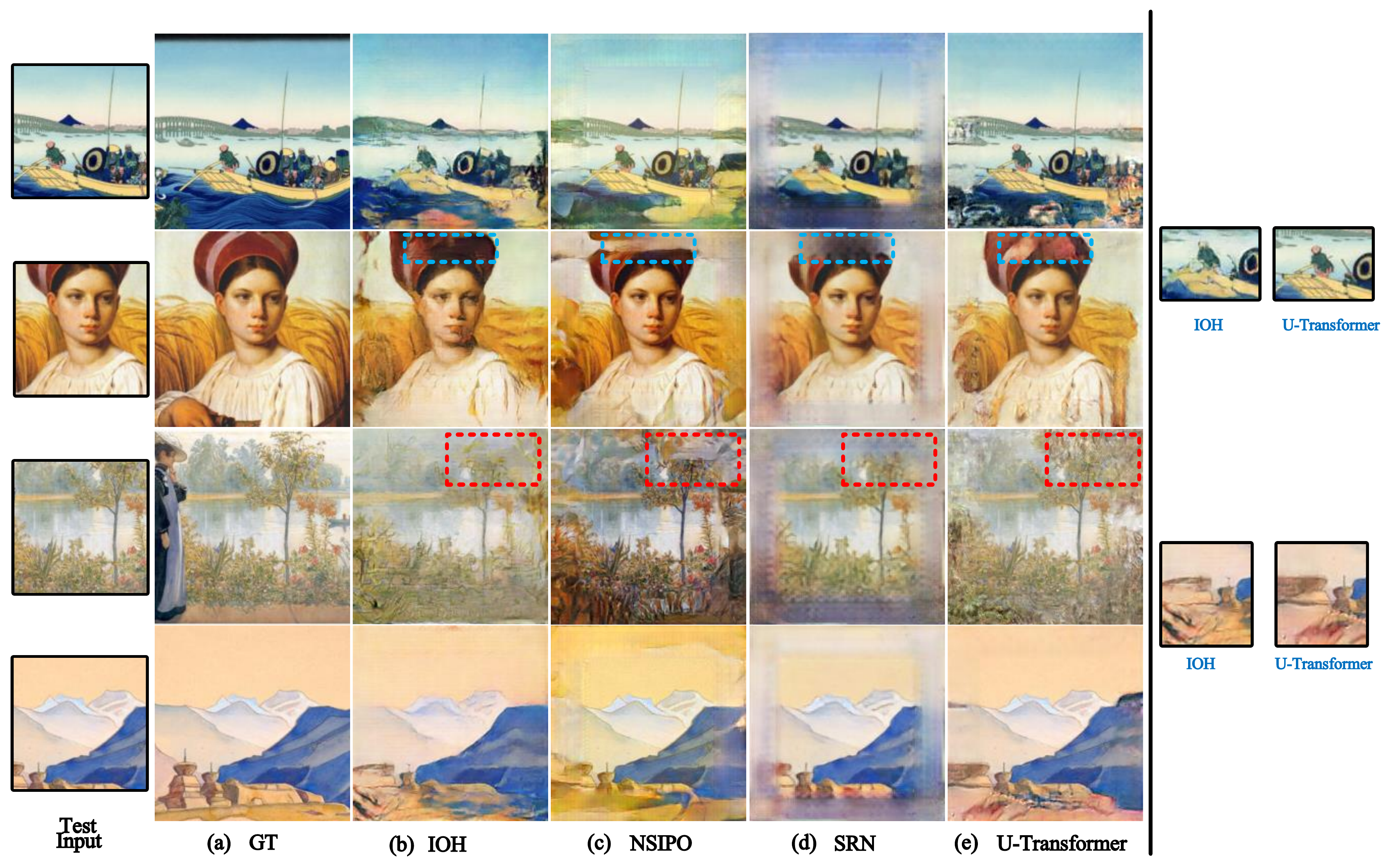}
    \caption{Qualitative results compared with other outpainting methods on wikiart dataset.}
    \label{wikiart}
\end{figure}

\subsubsection{Qualitative Results}
Fig.~\ref{Quali}, Fig.~\ref{build} and Fig.~\ref{wikiart} show some examples of generalised outpainting results on the scenery, building, and wikiart datasets indicating that our proposed model generates visually appealing results, especially for the four corner regions.
The horizontal outpainting method NSIPO extrapolates one side of a given image, limiting generalised outpainting.
The BCT block in NSIPO is explicitly designed to predict feature maps horizontally, which expands features for all four sides respectively in generalised outpainting. The edges between the expanded part and the original image are apparent and the predicted regions tend to be distorted frequently.
Nevertheless, our model weakens the sense of edges between the original and generated pixels and makes the expanded part more realistic. Our model also achieves better results than SRN and IOH which are both CNN-based generalised image outpainting methods. They still suffer from blurriness and unsatisfying extrapolation in the generated images. Compared to IOH, our model could generate better-reconstructed windmills and rocks marked in the orange circle in columns (b) and (e) of Fig.~\ref{Quali}. Compared to NSIPO, the predicted regions of our model are more smooth and real with reasonable content in columns (c) and (e) of Fig.~\ref{build}. Moreover, the dotted box indicates that our model generates more realistic and seamless images in terms of details, e.g. the trees, rocks, and rivers predicted in column (e) of Fig.~\ref{Quali}.
The fine-art paintings are more complicated, having a broader variety of textures and styles. As illustrated in Fig.~\ref{wikiart}, the images generated by our U-Transformer are more predictable in terms of texture and stylistic consistency across the image, particularly in the extrapolate area.

\begin{table}[!htbp]
\caption{Ablation study of each component in our proposed model. w/o Trans: Replacing the transformer structure with CNN backbone. w/o SC\&TSP: Removing both the SC and TSP modules. w/o SC: Removing the U-Net skip connection. w/o TSP: Removing the TSP module.}
\label{qabla}
\centering
\begin{tabular}{c|cccc}
\hline
           & FID$\downarrow$    & IS$\uparrow$    & PSNR$\uparrow$   & SSIM$\uparrow$  \\ \hline
w/o Trans    & 37.934 & 2.991 & 22.658 & 0.743 \\
w/o SC\&TSP    & 29.390  & 3.092 & 23.333  & 0.768 \\
w/o SC        & 92.038 & 2.345 & 20.895 & 0.554 \\
w/o TSP     & 25.005 & 3.215 & 23.451 & 0.782 \\
Full model & 24.860 & 3.554 & 23.273  & 0.788 \\ \hline
\end{tabular}
\end{table}

\begin{figure}[!htbp]
    \centering
    \includegraphics[width=0.95\textwidth]{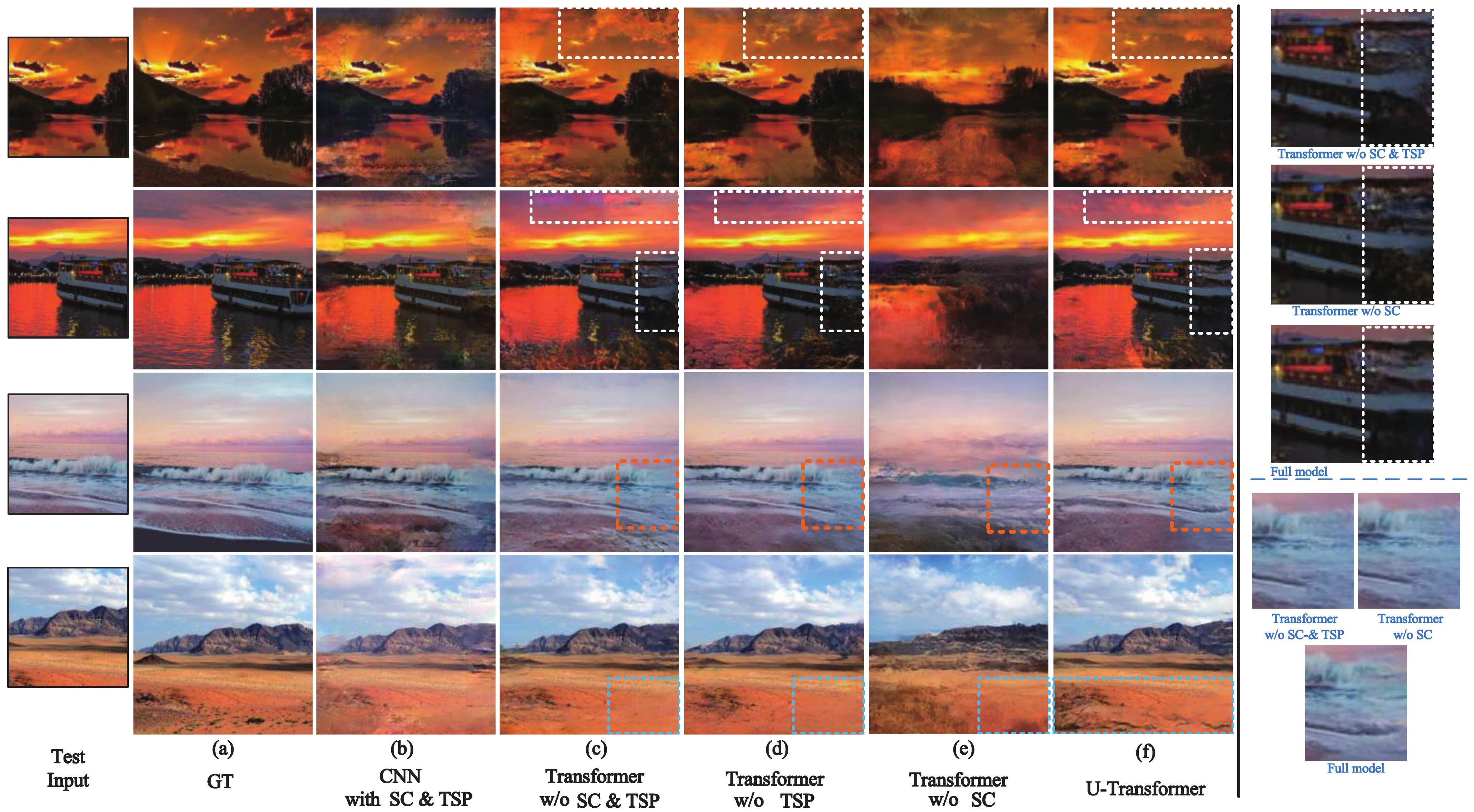}
    \caption{Visual results of ablation study on Scenery dataset.}
    \label{abl}
\end{figure}
\subsection{Ablation Study}
We performed an ablation study to demonstrate the necessity of each component in the proposed U-Transformer. Table~\ref{qabla} shows the quantitative results with different model variants where they are trained with the same training strategy on the scenery dataset having one specific component removed. We could see that removing any design of our full model leads to worse performance, proving that our designs could boost the proposed model learning and improve the quality of the generated extrapolation images. We further study the effectiveness of our designs from qualitative results, with some examples shown in Fig.~\ref{abl}.

Although removing the TSP module seems to achieve similar quantitative results to our full model, the image quality of the predicted ambient region is unsatisfactory. On the contrary, our full model with introducing the TSP module generates more delicate results with rich texture and exquisite details. The potentially temporal and spatial information could help the model predict more reasonable and coherent content. For example, in the third row of Fig.~\ref{abl}, our full model generates the whole waves smoothly and seamlessly with more consistent details in column (f) compared to other variants having broken waves marked in the orange box. Besides, the beach looks more realistic as well.

Overall, replacing transformer structure with CNN backbone in the variant of CNN with SC\&TSP suffers from blur and has an obvious boundary between the input and predicted region due to long-range dependencies not being well extracted.
When only using Swin Transformer blocks, the extended image borders have plausible structure but obvious blunt colours in column (c). When further utilizing the SC, more realistic and smooth colours are generated in the extended areas but not satisfactory details in column (d). While not adding SC but a TSP as the bottleneck, there is blurriness but generates vivid details in column (e). In general, only when employing both SC and TSP in the framework, could the full model be able to extrapolate around images with plausible structure and vivid details as shown in highlighted regions.

\begin{figure}[!htbp]
    \centering
    \includegraphics[width=0.6\textwidth]{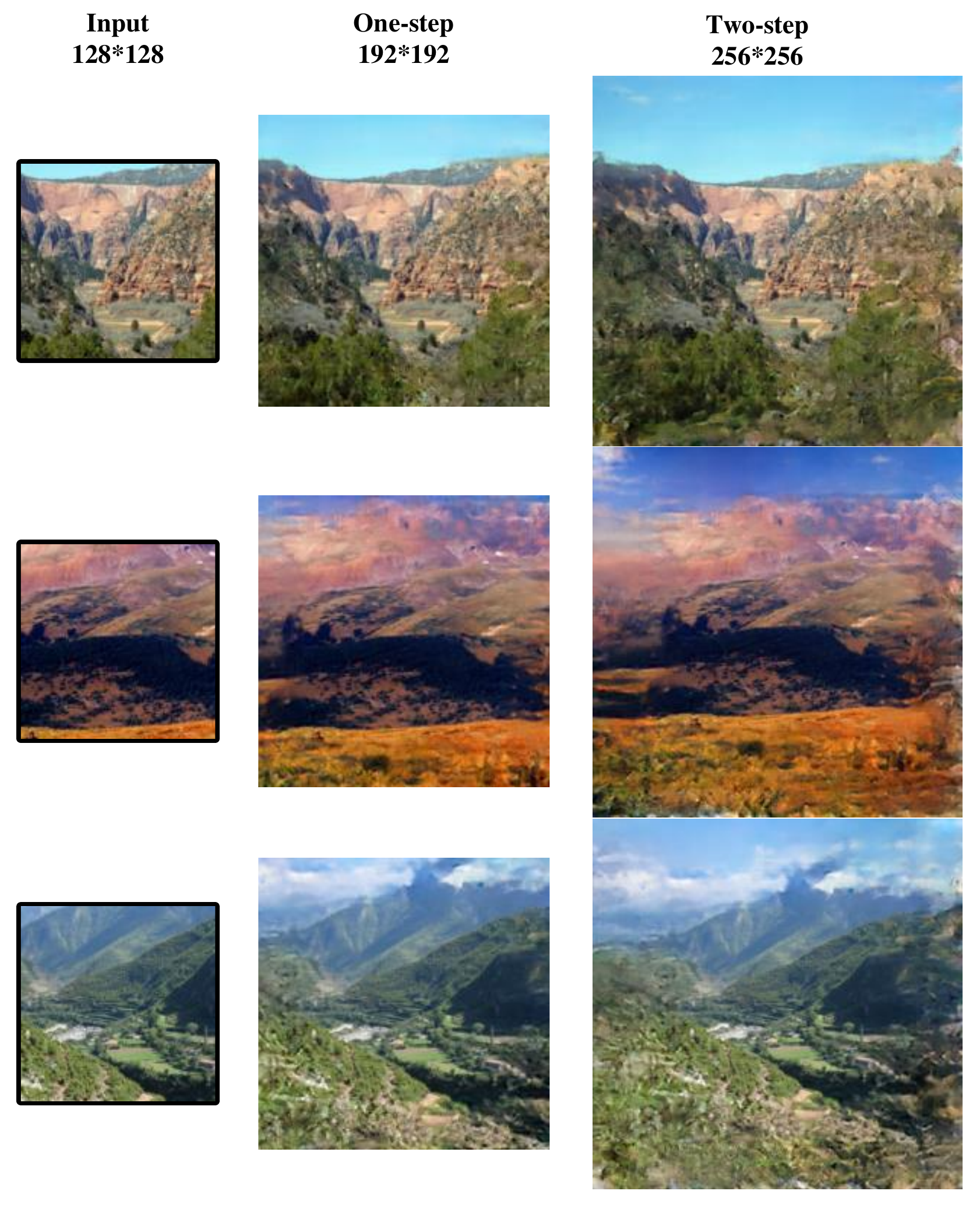}
    \caption{Visual results of two-step extrapolation. The first column represents the input sub-image with size $128\times 128$. The second column shows the one-step extrapolation with image size $192\times 192$. The third column shows the two-step extrapolation with image size $256\times 256$.}
    \label{two-step}
\end{figure}
\subsection{Visualization of two-step extrapolation}
We further conduct the two-step extrapolation experiments to demonstrate the effectiveness and superiority of the proposed TSP module.
In the testing stage, we first generate the one-step extrapolated images with the size of $192\times 192$. Then we adjust the predicting step in the TSP module and generate the two-step extrapolated images with the size of $256\times 256$ by using the output of one-step predicting as the input.
Our model could generate the arbitrary size of extrapolated images, although it has been trained with a fixed image size of $192\times 192$.
Some visual examples could be seen in Fig.~\ref{two-step}. These results show that by adjusting the predicting step our model could generate multi-step extrapolated images with semantic consistency and realistic contents.

\section{Conclusion \& Future Work}
In this paper, we have proposed the transformer-based generative adversarial network (U-Transformer) to improve the generalised image outpainting performance.
The Swin Transformer blocks overcame the intrinsic locality and captured image long-range dependencies well. Moreover, the Skip Connection and multi-view Temporal Spatial Predictor modules reinforced image self-reconstruction as well as unknown-part prediction smoothly and realistically.
The TSP module makes it tractable to handle feature maps with a larger masked part in the testing stage to generate multi-step extrapolated images.
Extensive experiments on scenery and building datasets proved the effectiveness of our methods.

Trained with a fixed resolution, our model could theoretically extrapolate images with the arbitrary resolution given an input sub-image in the testing stage. The generated images will contain more damaging information when the extrapolated images are larger. In the future, we will investigate to improve the Swin Transformer block making it preferable to generate high-quality images.



\bibliographystyle{elsarticle-num}
\bibliography{ref}






\end{document}